\documentclass[runningheads]{llncs}
\usepackage{graphicx}
\usepackage{comment}
\usepackage{amsmath,amssymb}
\usepackage{color}
\usepackage{makecell}
\usepackage{amsmath,amsfonts,bm}

\def\eqref#1{equation~\ref{#1}}
\def\1{\bm{1}}

\newcommand{\test}{\mathcal{D_{\mathrm{test}}}}

\def\vx{{\bm{x}}}

\DeclareMathAlphabet{\mathsfit}{\encodingdefault}{\sfdefault}{m}{sl}
\SetMathAlphabet{\mathsfit}{bold}{\encodingdefault}{\sfdefault}{bx}{n}

\def\gX{{\mathcal{X}}}
\def\gY{{\mathcal{Y}}}

\usepackage{epsfig}
\usepackage{xr}
\usepackage{multirow}
\usepackage{mathtools}
\usepackage{algorithmic}
\usepackage{float}
\usepackage[ruled, linesnumbered]{algorithm2e}
\usepackage[usenames,dvipsnames]{xcolor}
\usepackage{caption}
\usepackage{subcaption}
\usepackage{soul}
\usepackage{wrapfig}
\usepackage{hyperref}  %hyperref still needs to be put at the end!

\definecolor{amber(sae/ece)}{rgb}{1.0, 0.49, 0.0}
\definecolor{tangelo}{rgb}{0.98, 0.3, 0.0}
\definecolor{applegreen}{rgb}{0.0, 0.5, 0.0}

\def\cartoonratio{47}
\def\cartoondenseratio{42}
\def\cartoonsize{0.18}
\def\cartoondensesize{0.80}
\def\cartoonimgwidth{0.15}
\def\cyclgeganimgwidth{0.16}
\def\teaserwidth{0.23}
\def\ablaimgwidth{0.15}
\def\eg{\textit{e}.\textit{g}.}
\def\ie{\textit{i}.\textit{e}.}
\def\exp{{\mathbb{E}}}

\begin{document}
\pagestyle{headings}
\mainmatter
\def\ECCVSubNumber{1488}  % Insert your submission number here

\title{GAN Slimming: All-in-One GAN Compression by A Unified Optimization Framework} % Replace with your title

\begin{comment}
\titlerunning{ECCV-20 submission ID \ECCVSubNumber} 
\authorrunning{ECCV-20 submission ID \ECCVSubNumber} 
\author{Anonymous ECCV submission}
\institute{Paper ID \ECCVSubNumber}
\end{comment}
%******************

% \begin{comment}
\titlerunning{GAN Slimming: A Unified Optimization Framework for GAN Compression}
\author{Haotao Wang\inst{1} \and
Shupeng Gui\inst{2} \and
Haichuan Yang\inst{2} \\
Ji Liu\inst{3} \and
Zhangyang Wang\inst{1}
}
\authorrunning{H. Wang, S. Gui, H. Yang, J. Liu and Z. Wang}
\institute{
University of Texas at Austin, Austin TX 78712, USA \\
\email{\{htwang, atlaswang\}@utexas.edu}\\
\and
University of Rochester, Rochester NY 14627, USA\\
\email{\{sgui2, hyang36\}@ur.rochester.edu}
\and
AI Platform, Ytech Seattle AI Lab, FeDA Lab, Kwai Inc., Seattle WA 98004, USA\\
\email{ji.liu.uwisc@gmail.com}
}
% \end{comment}
%******************
\maketitle

\begin{abstract}
Generative adversarial networks (GANs) have gained increasing popularity in various computer vision applications, and recently start to be deployed to resource-constrained mobile devices. Similar to other deep models, state-of-the-art GANs suffer from high parameter complexities. That has recently motivated the exploration of compressing GANs (usually generators). Compared to the vast literature and prevailing success in compressing deep classifiers, the study of GAN compression remains in its infancy, so far leveraging individual compression techniques instead of more sophisticated combinations. We observe that due to the notorious instability of training GANs, heuristically stacking different compression techniques will result in unsatisfactory results. 
To this end, we propose the first unified optimization framework combining multiple compression means for GAN compression, dubbed \textbf{GAN Slimming} (GS). 
GS seamlessly integrates three mainstream compression techniques: model distillation, channel pruning and quantization, together with the GAN minimax objective, into one unified optimization form, that can be efficiently optimized from end to end. 
Without bells and whistles, GS largely outperforms existing options in compressing image-to-image translation GANs. Specifically, we apply GS to compress CartoonGAN, a state-of-the-art style transfer network, by up to $\mathbf{\cartoonratio \times}$ times, with minimal visual quality degradation. Codes and pre-trained models can be found at \url{https://github.com/TAMU-VITA/GAN-Slimming}.
\end{abstract}

\begin{figure}[ht] 
	\centering
	\setlength{\tabcolsep}{1pt}
	\begin{tabular}{ccccc}
    	Source image & \makecell{Original result\\56.46 GFLOPs\\42.34 MB} & \makecell{Our result (32bit)\\ {1.34 GFLOPs}\\\cartoondensesize~MB} & \makecell{Our result (8bit)\\ {1.20 GFLOPs}\\\cartoonsize~MB} \\
    	\includegraphics[width=\teaserwidth\linewidth]{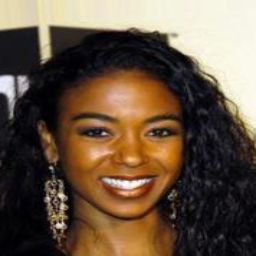} & 
		\includegraphics[width=\teaserwidth\linewidth]{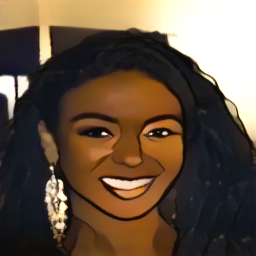} & 
		\includegraphics[width=\teaserwidth\linewidth]{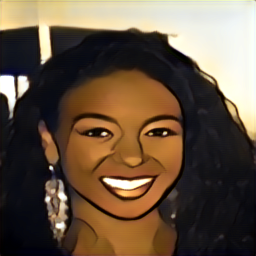} & 
		\includegraphics[width=\teaserwidth\linewidth]{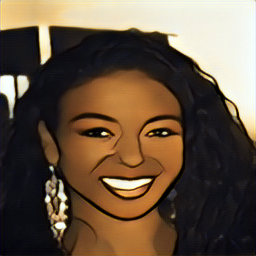}
		\\
		\includegraphics[width=\teaserwidth\linewidth]{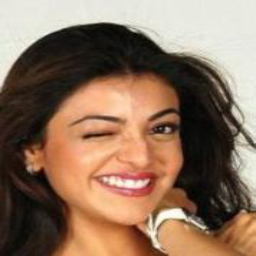} & 
		\includegraphics[width=\teaserwidth\linewidth]{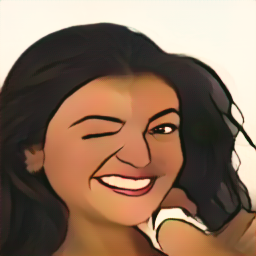} & 
		\includegraphics[width=\teaserwidth\linewidth]{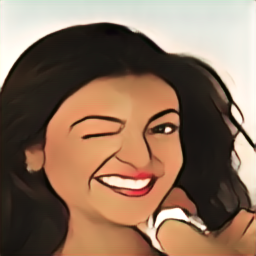} & 
		\includegraphics[width=\teaserwidth\linewidth]{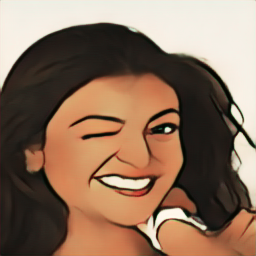} 
		\\
	\end{tabular}
	\caption{Representative visual examples by GAN Slimming on CartoonGAN \cite{chen2018cartoongan}.}
	\label{fig:teaser}
\end{figure}

\section{Introduction} \label{sec:intro}
Generative adversarial networks (GANs)~\cite{goodfellow2014generative}, especially, image-to-image translation GANs, have been successfully applied to image synthesis \cite{karras2019style}, style transfer \cite{chen2018cartoongan}, image editing and enhancement \cite{ledig2017photo,kupyn2019deblurgan,jiang2019enlightengan}, to name just a few. Due to the growing usage, there has been an increasing demand to deploy them on resource-constrained devices \cite{liu2020adadeep}. For example, many filter-based image editing applications now desire to run image-to-image translation GANs locally.
However, GANs, just like most other deep learning models, bear explosive parameter amounts and computational complexities. For example, in order to process a $256 \times 256$ image, a state-of-the-art style transfer network, CartoonGAN \cite{chen2018cartoongan}, would cost over 56 GFLOPs. Launching such models on mobile devices requires considerable memory and computation costs, which would be infeasible for most devices, or at least degrades user experience due to the significant latency. 

Existing deep model compression methods mainly focus on image classification or segmentation tasks, and were not directly applicable on GAN compression tasks due to notorious instability of GAN minimax training. For example, \cite{shu2019co} shows that generators compressed by state-of-the-art classifier compression methods \cite{hu2016network,liu2017learning,luo2017thinet} all suffer great performance decay compared with the original generator.
Combining (either heuristically cascading or jointly training) multiple different compression techniques, such as channel pruning, model distillation, quantization and weight sharing, has been shown to outperform separately using single compression techniques alone in traditional classification tasks \cite{wang2018graph,mishra2017apprentice,polino2018model}.  
In comparison, current methods \cite{shu2019co,chen2020distilling} have so far only tried to apply one single technique to compressing GANs.
\cite{shu2019co} proposed the first dedicated GAN compression algorithm: an evolutionary method based \underline{channel pruning} algorithm. However, the method is specifically designed for CycleGAN and non-straightforward to extend to GANs without cycle consistency structure (\eg, encoder-decoder GANs \cite{chen2018cartoongan,sanakoyeu2018style} that are also popular).
A latest work \cite{chen2020distilling} proposed to training an efficient generator by \underline{model distillation}. 
By removing the dependency on cycle consistency structure, \cite{chen2020distilling} achieves more general-purpose GAN compression than \cite{shu2019co}. However, the student network in \cite{chen2020distilling} is still hand-crafted and relies on significant architecture engineering for good performance.

As discussed in \cite{shu2019co,chen2020distilling}, applying a single compression technique to GANs is already challenging due to their notorious training instability. As one may imagine, integrating multiple compression techniques together for GAN compression will only further amplify the instability, putting an open question:
\begin{center}
\textit{Can we gain more from combining multiple compression means for GANs?\\ If yes, how to overcome the arising challenge of GAN instability?
}
\end{center}

Our answer is by presenting the first end-to-end optimization framework combining multiple compression means for general GAN compression, named \textbf{GAN Slimming (GS)}. 
The core contribution of GS is a unified optimization form, that seamlessly integrates three popular model compression techniques (channel pruning, quantization and model distillation), to be jointly optimized in a minimax optimization framework. GS pioneers to advance GAN compression into jointly leveraging multiple model compression methods, and demonstrate the feasibility and promise of doing so, despite the GAN instability. 

Experiments demonstrate that GS overwhelms state-of-the-art GAN compression options that rely on single compression means. For example, we compress the heavily-parameterized CartoonGAN by up to $\cartoonratio \times$, achieving nearly real-time cartoon style transfer on mobile devices, with minimal visual quality loss. Moreover, we have included a detailed ablation study for a deeper understanding of GS. Specifically, we demonstrate that \textit{naively stacking different compression techniques cannot achieve satisfactory GAN compression, sometimes even hurting catastrophically}, therefore testifying the necessity of our unified optimization. We also verify the effectiveness of incorporating the minimax objective into this specific problem.

% Our contributions are summarized in the following folds:
% \begin{itemize}
%     \item We are the first one to push forward GAN compression into jointly leveraging multiple compression techniques, and demonstrate its feasibility despite the GAN instability. 
%     \item We propose a unified optimization framework that integrate three popular model compression techniques and the GAN minimax objective, that can be end-to-end optimized.
%     \item Our method's performance overwhelms existing GAN compression options. Specifically, we compress the heavily-parameterized CartoonGAN by up to $\cartoonratio \times$, achieving nearly real-time cartoon style transfer on mobile devices, with minimal visual quality loss.
%     \item We present a detailed ablation study for a deeper understanding of GS. Specifically, we demonstrate the success of GS to arise from unified optimization rather than simply stacking multiple techniques. We also show the effectiveness of incorporating minimax objective into this specific problem.    
% \end{itemize}

\section{Related Works}

\subsection{Deep Model Compression}
Many model compression methods, including knowledge distillation~\cite{polino2018model}, pruning~\cite{han2015deep,he2017channel} and quantization~\cite{han2015deep}, have been investigated to compress large deep learning models, primarily classifiers~\cite{chen2020frequency,chen2020learning,chen2020addernet,han2020ghostnet,shen2019searching,singh2020leveraging,wang2018learning,wang2018packing,wu2018deep}. Structured pruning~\cite{wen2016learning}, such as channel pruning \cite{huang2018data,liu2017learning,molchanov2019importance,wen2016learning,yang2019ecc}, result in hardware-friendly compressed models and thus are widely adopted in real-world applications.
The authors of \cite{wen2016learning} enforced structured sparsity constraint on each layer's kernel weights, aided by group Lasso \cite{kim2010tree} to solve the optimization. \cite{liu2017learning} added $\ell_1$ constraint on the learnable scale parameters of batch normalization layers in order to encourage channel sparsity, and used subgradient descent to optimize the $\ell_1$ loss. Similarly, \cite{huang2018data} also utilized sparsity constraint on channel-wise scale parameters, solved with an accelerated proximal gradient method \cite{parikh2014proximal}. 

Quantization, as another popular compression means, reduces the bit width of the element-level numerical representations of weights and activations. Earlier works \cite{han2015deep,lin2017runtime,zhu2016trained} presented to quantize all layer-wise weights and activations to the same low bit width, \eg, from 32 bits to 8 bits or less. The model could even consist of only binary weights in the extreme case~\cite{courbariaux2015binaryconnect,rastegari2016xnor}. Note that, introducing quantization into network weights or activations will result in notable difficulty for propagating gradients \cite{hubara2017quantized}. Straight-through estimator (STE) \cite{courbariaux2015binaryconnect} is a successful tool to solve this problem by a proxy gradient for back propagation.

Knowledge distillation was first developed in \cite{Hinton_2015} to transfer the knowledge in an ensemble of models to a single model, using a soft target distribution produced by the former models. It was later on widely used to obtain a smaller network (student model), by fitting the ``soft labels'' (probabilistic outputs) generated from a trained larger network (teacher model). \cite{bulo2016dropout} used distillation to train a more efficient and accurate predictor. \cite{lopez_2015} unified distillation and privileged information into one generalized distillation framework to learn better representations.
\cite{wang2018adversarial,chen2019data} used generative adversarial training for model distillation.

\subsubsection{Combination of multiple compression techniques}
For compressing a deep classifier, \cite{tung2018clip,yang2019controllable} proposed to jointly train (unstructured) pruning and quantization together. \cite{wang2018graph,theis2018faster} adopted knowledge distillation to fine-tune a pruned student network, by utilizing the original dense network as teacher, which essentially followed a two-step cascade pipeline. Similarly, \cite{mishra2017apprentice,polino2018model} used full-precision networks as teachers to distill low-precision student networks. 
\cite{gui2019model} showed jointly training pruning and quantization can obtain compact classifiers with state-of-the-art trade-off between model efficiency and adversarial robustness.

Up to our best knowledge, all above methods cascade or unify two compression techniques, besides that they investigate compressing deep classifiers only. In comparison, our proposed framework jointly optimize three methods in one unified form\footnote{A concurrent work \cite{zhao2020smartexchange} jointly optimized pruning, decomposition, and quantization, into one unified framework for reducing the memory storage/access.}, that is innovative even for general model compression. It is further adapted for the special GAN scenario, by incorporating the minimax loss.

\subsection{GAN Compression}

GANs have been successful on many image generation and translation tasks \cite{goodfellow2014generative,gui2020review,guo2020positive,karras2017progressive,miyato2018spectral}, yet their training remains notoriously unstable. Numerous techniques were developed to stabilize the GAN training, \eg, spectral normalization \cite{miyato2018spectral}, gradient penalty \cite{gulrajani2017improved} and progressive training \cite{karras2017progressive}. 
%To the best of our knowledge, \cite{shu2019co,chen2020distilling} are so far the only two published algorithm specifically discussing GAN compression. 
As discussed in \cite{shu2019co,chen2020distilling}, the training difficulty causes extra challenges for compressing GANs, and failed many traditional pruning methods for classifiers such as \cite{luo2017thinet,hu2016network,liu2017learning}.

The authors of \cite{shu2019co} proposed the first dedicated GAN compression method: a co-evolution algorithm based channel pruning method for CycleGAN. Albeit successfully demonstrated on the style transfer application, their method faces several limitations.
First, their co-evolution algorithm relies on the cycle consistency loss to simultaneously compress generators of both directions. It is hence non-straightforward to extend to image-to-image GANs without cycle consistent loss (\eg, encoder-decoder GANs \cite{chen2018cartoongan,sanakoyeu2018style}). 
Second, in order to avoid the instability in GAN training, the authors model GAN compression as a ``dense prediction'' process by fixing the original discriminator instead of jointly updating it with the generator in a minimax optimization framework. This surrogate leads to degraded performance of the compressed generator, since the fixed discriminator may not suit the changed (compressed) generator capacity.
These limitations hurdle both its broader application scope and performance. 

The latest concurrent work \cite{chen2020distilling} explored model distillation: to guide the student to effectively inherit knowledge from the teacher, the authors proposed to jointly distill generator and discriminator in a minimax two-player game. \cite{chen2020distilling} improved over \cite{shu2019co} by removing the above two mentioned hurdles. However, as we observe from experiments (and also confirmed with their authors), the success of \cite{chen2020distilling} hinges notably on the appropriate design of student network architectures. Our method could be considered as another important step over \cite{chen2020distilling}, that ``learns'' the student architecture jointly with the distillation, via pruning and quantization, as to be explained by the end of Section 3.1.

%it is still using hand-crafted student network structures which require significant architecture engineering to achieve good performance. Our method differs from \cite{shu2019co,chen2020distilling} by jointly considering three different compression techniques in one optimization framework.

\section{The GAN Slimming Framework}

Considering a dense full-precision generator $G_0$ which converts the images from one domain $\gX$ to another $\gY$, our aim is to obtain a more efficient generator $G$ from $G_0$, such that their generated images $\{G_0(x), x \in \gX\}$ and $\{G(x), x \in \gX\}$ have similar style transfer qualities. 
In this section, we first outline the unified optimization form of our GS framework combining model distillation, channel pruning and quantization~(Section~\ref{sec:overall-framework}). We then show how to solve each part of the optimization problem respectively~(Section \ref{sec:optimization}), and eventually present the overall algorithm~(Section \ref{sec:implementation}).

\subsection{The Unified Optimization Form} \label{sec:overall-framework}
% image sets $\{x_i\}_{i=1}^{N} \in \gX$ and $\{y_i\}_{i=1}^{M} \in \gY$

We start formulating our GS objective from the traditional minimax optimization problem in GAN:
\begin{equation} \label{eq:GAN}
    \min_{G}\max_{D} L_{GAN},~\text{where } L_{GAN}=\exp_{y\in\gY}[\text{log}(D(y))] + \exp_{x\in\gX}[\text{log}(1-D(G(x))],
\end{equation}
where $D$ is the discriminator jointly trained with efficient generator $G$ by minimax optimization. Since $G$ is the functional part to be deployed on mobile devices and $D$ can be discarded after training, we do not need to compress $D$.  
Inspired by the success of model distillation in previous works~\cite{Hinton_2015,chen2020distilling}, we add a model distillation loss term $L_{dist}$ to enforce the small generator $G$ to mimic the behaviour of original large generator $G_0$, where $\textrm{d}(\cdot,\cdot)$ is some distance metric:
\begin{equation} \label{eq:distillation}
    L_{dist} = \exp_{x\in\gX} [\textrm{d}(G(x), G_0(x))],
\end{equation}
The remaining key question is: how to properly define the architecture of $G$? Previous methods \cite{Hinton_2015,chen2020distilling} first hand-crafted the smaller student model's architecture and then performed distillation. However, it is well known that the choice of the student network structure will affect the final performance notably too, in addition to the teacher model's strength. 

Unlike existing distillation methods \cite{chen2020distilling}, we propose to \textit{jointly infer} the $G$ architecture together with the distillation process. Specifically, we assume that $G$ can be ``slimmed'' from $G_0$, through two popular compression operations: channel pruning and quantization. For channel pruning, we follow \cite{liu2017learning} to apply $L_1$ norm on the trainable scale parameters $\gamma$ in the normalization layers to encourage channel sparsity: $L_{cp} = \|\gamma\|_1$. Denoting all other trainable weights in $G$ as $W$, we could incorporate the channel pruning via such sparsity constraint into the distillation loss in Eq.~(\ref{eq:distillation}) as below:
\begin{align} 
\begin{split} \label{eq:cp}
    L_{dist}(W,\gamma) + \rho L_{cp}(\gamma)
    = \exp_{x\in\gX} [\textrm{d}(G(x;W,\gamma), G_0(x))] + \rho \|\gamma\|_1,
\end{split}    
\end{align}
where $\rho$ is the trade-off parameter controlling the network sparsity level. Further, to integrate quantization, we propose to quantize both activations and weights,\footnote{We only quantize $W$, while always leaving $\gamma$ unquantized.} using two quantizers $q_a(\cdot)$ and $q_w(\cdot)$, respectively, to enable the potential flexibility for hybrid quantization \cite{wang2019haq}. While it is completely feasible to adopt learnable quantization intervals \cite{jung2019learning}, we adopt uniform quantizers with pre-defined bit-width for $q_a(\cdot)$ and $q_w(\cdot)$, respectively, for the sake of simplicity (including hardware implementation ease). The quantized weights can be expressed as $q_w(W)$, while we use $G_q$ to denote generators equipped with activation quantization $q_a(\cdot)$ for notation compactness. Eventually, the final objective combining model distillation, channel pruning and quantization has the following form:
\begin{align} 
\begin{split} \label{eq:overall}
    L(W,\gamma,\theta) &= L_{GAN}(W,\gamma,\theta) + \beta L_{dist}(W,\gamma) + \rho L_{cp}(\gamma)\\
    &= 
    \exp_{y\in\gY}~[\text{log}(D(y;\theta))] 
    + 
    \exp_{x\in\gX}~[
    \text{log}(1-D({\color{red}G_q}(x;{\color{red}q_w}(W),\gamma);\theta))] \\
    &+ 
    \exp_{x\in\gX}~[{\color{blue}\beta\textrm{d}(}{\color{red}G_q}(x;{\color{red}q_w}(W),\gamma), G_0(x){\color{blue})}] \\
    &+ 
    {\color{applegreen}\rho\|\gamma\|_1},
\end{split}    
\end{align}
where $\theta$ represents the parameters in $D$.
The {\color{blue}blue} parts represent the distillation component, {\color{applegreen}green} represents channel pruning {\color{red}red} represents quantization. The above Eq.~(\ref{eq:overall}) is the target objective of GS, which is to be solved in a minimax optimization framework: 
\begin{align} 
\begin{split} \label{eq:opt}
\min_{W,\gamma}\max_{\theta}~L(W,\gamma,\theta)
\end{split}    
\end{align}

\paragraph{Connection to AutoML Compression.} Our framework could be alternatively interpreted as performing a special neural architecture search (NAS) \cite{he2018amc,gong2019autogan} to obtain the student model, where the student's architecture needs be ``morphable'' from the teacher's through only pruning and quantization operations. 
% One related work  \cite{bashivan2019teacher} demonstrated that using a teacher model's guidance helps search for more compact models in NAS.

Interestingly, two concurrent works \cite{fu2020autogan,li2020gan} have successfully applied NAS to search efficient generator architectures, and both achieved very promising performance too. We notice that a notable portion of the performance gains shall be attributed to the carefully designed search spaces, as well as computationally intensive search algorithms. In comparison, our framework is based on an end-to-end optimization formulation, that (1) has explainable and well-understood behaviors; (2) is lighter and more stable to solve; and (3) is also free of the NAS algorithm’s typical engineering overhead (such as defining the search space and tuning search algorithms). Since our method directly shrinks the original dense model via pruning and quantization only, it cannot introduce any new operator not existing in the original model. That inspires us to combine the two streams of compression ideas (optimization-based versus NAS-based), as future work.

% our GS is enjoys two main advantages. First, GS is accomplished through an end-to-end analytical optimization. It can jointly learn the student model's architecture and weights in one run, and does not rely on the (more tedious) sampling process in a given architecture search space. 
% Second, our GS can explore more fine-grained architecture spaces (\eg, adjusting channel numbers continuously) compared with NAS based methods, where the channel numbers in the searching space are usually several pre-defined discrete values.
% % It would be our future interest to compare the achievable performance with $G$ (jointly) obtained by NAS.
% Moreover, it is also straightforward to combine our GS with efficient network architecture designs such as depth-wise convolutional blocks \cite{howard2017mobilenets,sandler2018mobilenetv2}, as used in search spaces of \cite{fu2020autogan,li2020gan}, to further reduce computational costs and model sizes while keeping generation quality.
% }

\subsection{End-to-End Optimization} \label{sec:optimization}
The difficulties of optimizing~(\ref{eq:opt}) can be summarized in three-folds.
First, the minimax optimization problem itself is unstable. Second, updating $W$ involves non-differentiable quantization operations. Third, updating $\gamma$ involves a sparse loss term that is also non-differentiable. Below we discuss how to optimize them.

\subsubsection{Updating $W$} \label{sec:W-step}
The sub-problem for updating $W$ in Eq.~(\ref{eq:opt}) is:
\begin{align} 
\begin{split} \label{eq:W_step}
    &\min_{W}~L_W(W), \\
    &\text{where}~L_W(W) = \exp_{x\in\gX}[\text{log}(1-D(G_q(x;q_w(W),\gamma);\theta)) \\
    & \hspace{10em} + \beta \textrm{d}(G_q(x;q_w(W),\gamma), G_0(x))],
\end{split}    
\end{align}
To solve (\ref{eq:W_step}) with gradient-based methods, we need to calculate $\nabla_W L_W$, which is difficult due to the non-differentiable $q_a(\cdot)$ and $q_w(\cdot)$.
We now define the concrete form of $q_a(\cdot)$, $q_w(\cdot)$ and then demonstrate how to back propagate through them in order to calculate $\nabla_{W} L_W$.
Since both $q_a(\cdot)$ and $q_w(\cdot)$ are elementwise operations, we only discuss how they work on scalars.
We use $a$ and $w$ to denote a scalar element in the activation and convolution kernel tensors respectively.

When quantizing activations, we first clamp activations into range $[0, p]$ to bound the values, and then use $s_a={p / 2^m}$ as a scale factor to convert the floating point number to $m$ bits integers: $\text{round}(\min(\max(0, a), p) / s_a)$. Thus the activation quantization operator is as follows:
\begin{align}
q_a(a) = \text{round}(\min(\max(0, a), p) / s_a) \cdot s_a.
\end{align}
For weights quantization, we keep the range of the original weights and use symmetric coding for positive and negative ranges to quantize weights to $n$ bits. Specifically, the scale factor $s_w = {\|w\|_{\infty} / 2^{(n-1)}}$, leading to the quantization operator for weights:
\begin{align}
q_w(w) = \text{round}(w / s_w) \cdot s_w.
\end{align}
Since both quantization operators are non-differentiable, we use a proxy as the ``pseudo'' gradient in the backward pass, known as the straight through estimator (STE). For the activation quantization, we use
\begin{align}
{\partial q_a(a) \over \partial a} = 
\begin{cases}
1 \quad &\text{if $0 \leq a \leq p$};\\
0 &\text{otherwise}.
\end{cases}
\end{align}
Similarly for the weight quantization, the pseudo gradient is set to 
\begin{align}
    {\partial q_w(w) \over \partial w} = 1.
\end{align}
Now that we have defined the derivatives of $q_a(\cdot)$ and $q_w(\cdot)$, we can calculate $\nabla_W L_W$ through back propagation and update $W$ using the Adam optimizer~\cite{kingma2014adam}.

\subsubsection{Updating $\gamma$} \label{sec:gamma-step}
The sub-problem for updating $\gamma$ in Eq.~(\ref{eq:overall}) is a sparse optimization problem with a non-conventional fidelity term:
\begin{align} 
\begin{split} \label{eq:gamma_step}
    &\min_{\gamma} L_{\gamma}(\gamma) + \rho\|\gamma\|_1, \\
    &\text{where}~L_{\gamma}(\gamma) = \exp_{x\in\gX}[\text{log}(1-D(G_q(x;q_w(W),\gamma);\theta)) \\
    & \hspace{6em} + \beta \textrm{d}(G_q(x;q_w(W),\gamma), G_0(x))],
\end{split}
\end{align}
We use the proximal gradient to update $\gamma$ as follows:
\begin{align} 
    g_{\gamma}^{(t)} &\gets \nabla_{\gamma} L_{\gamma}(\gamma)\bigg\rvert_{\gamma=\gamma^{(t)}} \\
    \gamma^{(t+1)} &\gets \textrm{prox}_{\rho\eta^{(t)}}(\gamma^{(t)}-\eta^{(t)}g_{\gamma}^{(t)})
\end{align}
where $\gamma^{(t)}$ and $\eta^{(t)}$ are the values of $\gamma$ and learning rate at step $t$, respectively. The proximal function $\textrm{prox}_\lambda(\cdot)$ for the $\ell_1$ constraint is the soft threshold function:
\begin{align} 
    \textrm{prox}_\lambda(\vx) = \textrm{sgn}(\vx) \odot \textrm{max}(|\vx|-\lambda\mathbf{1}, \mathbf{0})
\end{align}
where $\odot$ is element-wise product, $\textrm{sgn}(\cdot)$ and $\textrm{max}(\cdot,\cdot)$ are element-wise sign and maximum functions respectively.

\subsubsection{Updating $\theta$}
The sub-problem of updating $\theta$ is the inner maximization problem in Eq.~(\ref{eq:opt}), which we solve by the gradient ascent method:
\begin{align} 
\begin{split} \label{eq:tyheta_step}
    &\max_{\theta}~L_\theta(\theta), \\
    &\text{where}~L_\theta(\theta) = \exp_{y\in\gY}[\text{log}(D(y;\theta))] + \exp_{x\in\gX}[\text{log}(1-D(G_q(x);\theta))],
\end{split}    
\end{align}
We iteratively update $D$ (parameterized by $\theta$) and $G$ (parameterized by $W$ and $\gamma$) following \cite{zhu2017unpaired}.

\subsection{Algorithm Implementation} \label{sec:implementation}

Equipped with the above gradient computation, the last missing piece in solving problem~(\ref{eq:overall}) is to choose $d$. Note that, most previous distillation works are for classification-type models with softmax outputs (soft labels), and therefore adopt KL divergence. For GAN compression, the goal of distillation shall minimize the discrepancy between two sets of generated images. To this end, we adopt the perceptual loss \cite{johnson2016perceptual} as our choice of $d$. It has shown to effectively measure not only low-level visual cue, but also high-level semantic differences between images, and has been popularly adopted to regularizing GAN-based image generation.

\begin{wrapfigure}[16]{R}{0.58\textwidth}
\begin{flushright}
\begin{minipage}{0.58\textwidth}
\begin{algorithm}[H] 
\SetAlgoLined
\KwIn{$\gX$, $\gY$, $\beta$, $\rho$, $T$, $\{\alpha^{(t)}\}_{t=1}^{T}, \{\eta^{(t)}\}_{t=1}^{T}$.
}
\KwOut{$W, \gamma$}  % no need to return \theta.
Random initialization: $W^{(1)}, \gamma^{(1)}, \theta^{(1)}$ \\
\For{$t \gets 1$ \KwTo $T$}
{
    Get a batch of data from $\gX$ and $\gY$; \\
    $W^{(t+1)} \gets W^{(t)} - \alpha^{(t)}\nabla_{W} L_W$; \\
    $\gamma^{(t+1)} \gets \textrm{prox}_{\rho\eta^{(t)}}(\gamma^{(t)}-\eta^{(t)}\nabla_{\gamma} L_{\gamma})$; \\
    $\theta^{(t+1)} \gets \theta^{(t)} + \alpha^{(t)}\nabla_{\theta} L_\theta$; \\
}
$W \gets q_w(W^{T+1})$ \\
$\gamma \gets \gamma^{T+1}$
\caption{GAN Slimming (GS)}
\label{alg:GS}
\end{algorithm}
\end{minipage}
\end{flushright}
\end{wrapfigure}

Finally, Algorithm~\ref{alg:GS} summarizes our GS algorithm with end-to-end optimization. By default, we quantize both activation and kernel weights uniformly to 8-bit (\ie, $m$ = $n$ = $8$) and set activation clamping threshold $p$ to $4$.
We use Adam ($\beta_1$ = $0.9$, $\beta_2$ = $0.5$, following \cite{zhu2017unpaired}) to update $W$ and $\theta$, and SGD to update $\gamma$. We also use two groups of learning rates $\alpha^{(t)}$ and $\eta^{(t)}$ for updating $\{W,\theta\}$ and $\gamma$ respectively. $\alpha^{(t)}$ starts to be decayed linearly to zero, from the $T/2$-th iteration, while $\eta^{(t)}$ is decayed using a cosine annealing scheduler.

\section{Experiments}

\subsection{Unpaired Image Translation with CycleGAN} \label{sec:CyleGAN}
Image translation and stylization is currently an important motivating application to deploy GANs on mobile devices. In this section, we compare GS with the only two published GAN compression methods CEC \cite{shu2019co} and GD \cite{chen2020distilling} on horse2zebra \cite{zhu2017unpaired} and summer2winter \cite{zhu2017unpaired} datasets. 
Following \cite{shu2019co}, we use model size and FLOPs to measure the efficiency of generator and use FID \cite{heusel2017gans} between source style test set transfer results and target style test set to quantitatively measure the effectiveness of style transfer. We used the same implementation of FID as \cite{shu2019co} for fair comparison.
The metric statistic of original CycleGAN is summarized in Table~\ref{tab:cyclegan_stat}. We denote the original dense model as $G_0$ and an arbitrary compressed generator as $G$. Following \cite{shu2019co}, we further define the following three metrics to evaluate efficiency-quality trade-off of different compression methods:
\[
    r_c = \frac{\textrm{ModelSize}_{G_0}}{\textrm{ModelSize}_{G}},
    ~r_s = \frac{\textrm{FLOPs}_{G_0}}{\textrm{FLOPs}_G},
    ~r_f = \frac{\textrm{FID}_{G_0}}{\textrm{FID}_G}.
\]
Larger $r_c$ and $r_s$ indicate more model compactness and efficiency and larger $r_f$ indicates better style transfer quality.

\begin{table}[ht] 
\begin{center}
\caption{Statistics of the original CycleGAN model: FLOPs, model size and FID on different tasks.}
\label{tab:cyclegan_stat}
\begin{tabular}{c|c|c|c|c|c}
\hline
\multirow{2}{*}{GFLOPs} & \multirow{2}{*}{\makecell{Memory\\(MB)}} & \multicolumn{4}{c}{FID}  \\
\cline{3-6} & & horse-to-zebra & zebra-to-horse & summer-to-winter & winter-to-summer \\
\hline
52.90 & 43.51 & 74.04 & 148.81 & 79.12 & 73.31 \\
\hline
\end{tabular}
\end{center}
\end{table}

% \begin{wrapfigure}[19]{R}{0.67\textwidth}
% \begin{flushright}
% \vspace{-4em}
% \begin{minipage}{0.67\textwidth}
% \begin{table}[H]
\begin{table}[ht]
\begin{center}
\caption{Compassion with the state-of-the-art GAN compression methods \cite{shu2019co} and \cite{chen2020distilling} on CycleGAN compression. The best metric is shown in bold and the second best is underlined.}
\label{tab:quant_sota_compare}
\begin{tabular}{c|c|c|c|c|c}
\hline
\multirow{2}{*}{Task} & \multirow{2}{*}{Metric} & \multicolumn{4}{c}{Method} \\
\cline{3-6} &  & CEC~\cite{shu2019co} & GD~\cite{chen2020distilling} & GS-32 & GS-8 \\
\hline\hline
\multirow{3}{*}{horse-to-zebra} 
    & $r_s$ & 4.23 & 3.91 & \underline{4.66} & \textbf{4.81} \\
    & $r_c$ & 4.27 & 4.00 & \underline{5.05} & \textbf{21.75} \\
    & $r_f$ & 0.77 & 0.76 & \textbf{0.86} & \underline{0.84} \\
\hline
\multirow{3}{*}{zebra-to-horse} 
    & $r_s$ & 4.35 & 3.91 & \underline{4.39} & \textbf{4.40} \\
    & $r_c$ & 4.34 & 4.00 & \underline{4.81} & \textbf{21.00} \\
    & $r_f$ & 0.94 & 0.99 & \underline{1.24} & \textbf{1.25} \\
\hline
\multirow{3}{*}{summer-to-winter} 
    & $r_s$ & 5.14 & 3.91 & \underline{6.21} & \textbf{7.18} \\
    & $r_c$ & 5.44 & 4.00 & \underline{6.77} & \textbf{38.10} \\
    & $r_f$ & 1.01 & 1.08 & \textbf{1.13} & \underline{1.12} \\
\hline
\multirow{3}{*}{winter-to-summer} 
    & $r_s$ & 5.17 & 3.91 & \underline{6.01} & \textbf{6.36} \\
    & $r_c$ & 5.70 & 4.00 & \underline{6.17} & \textbf{31.22} \\
    & $r_f$ & 0.93 & 0.97 & \underline{0.98} & \textbf{1.01} \\
\hline
\end{tabular}
\end{center}
\end{table}
% \end{minipage}
% \vspace{-3em}
% \end{flushright}
% \end{wrapfigure}

\begin{figure*}[ht] 
	\centering
	\setlength{\tabcolsep}{1pt}
	\begin{tabular}{cccccc}
		Source image & Original \cite{zhu2017unpaired} & CEC \cite{shu2019co} & GD \cite{chen2020distilling} & GS-32 & GS-8
		\\
		& \makecell{52.90 G\\43.51 MB} & \small \makecell{12.51 G\\10.19 MB} & \small \makecell{13.51 G\\10.86 MB} & \small \makecell{11.34 G\\8.61 MB} & \small \makecell{10.99 G\\2.15 MB} \\
        \includegraphics[width=\cyclgeganimgwidth\linewidth]{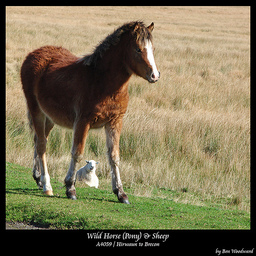} &
		\includegraphics[width=\cyclgeganimgwidth\linewidth]{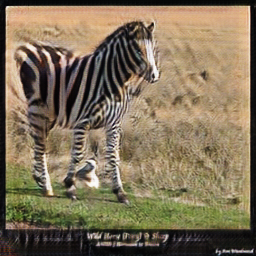} & 
		\includegraphics[width=\cyclgeganimgwidth\linewidth]{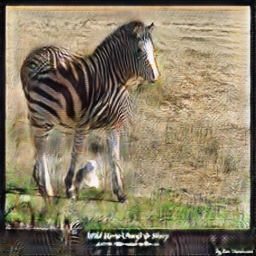} &
		\includegraphics[width=\cyclgeganimgwidth\linewidth]{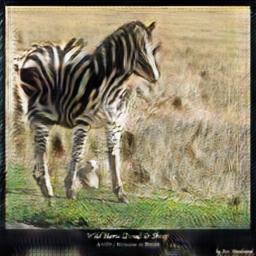} &
		\includegraphics[width=\cyclgeganimgwidth\linewidth]{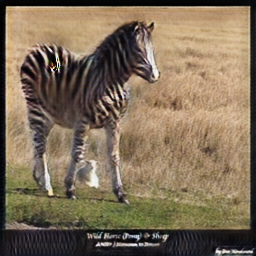} &
		\includegraphics[width=\cyclgeganimgwidth\linewidth]{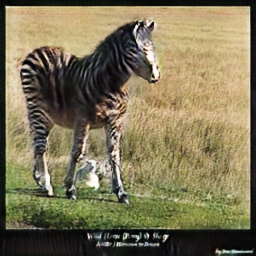}
		\\
		\includegraphics[width=\cyclgeganimgwidth\linewidth]{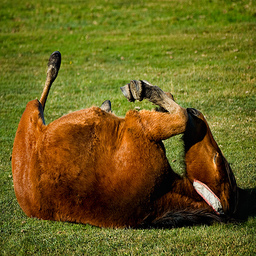} &
		\includegraphics[width=\cyclgeganimgwidth\linewidth]{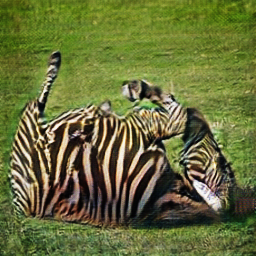} & 
		\includegraphics[width=\cyclgeganimgwidth\linewidth]{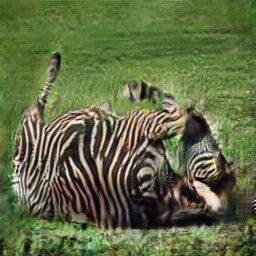} &
		\includegraphics[width=\cyclgeganimgwidth\linewidth]{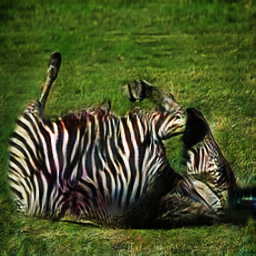} &
		\includegraphics[width=\cyclgeganimgwidth\linewidth]{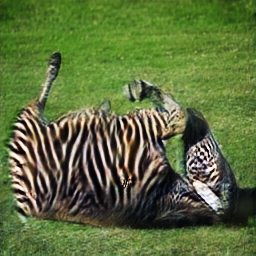} &
		\includegraphics[width=\cyclgeganimgwidth\linewidth]{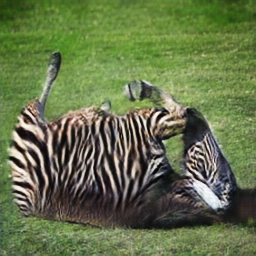}
		\\
		& \makecell{52.90 G\\43.51 MB} & \makecell{10.29 G\\10.18 MB} & \makecell{13.51 G\\10.86 MB} & \makecell{ 8.52 G\\1.61 MB} & \makecell{7.37 G\\1.14 MB} \\
		\includegraphics[width=\cyclgeganimgwidth\linewidth]{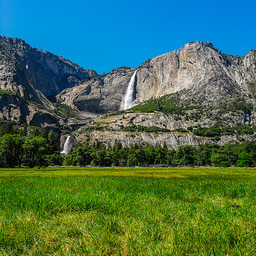} &
		\includegraphics[width=\cyclgeganimgwidth\linewidth]{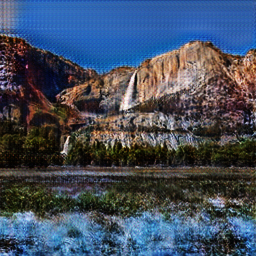} & 
		\includegraphics[width=\cyclgeganimgwidth\linewidth]{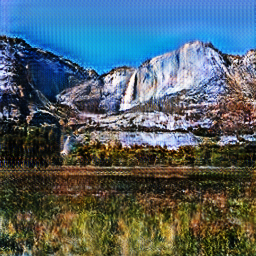} &
		\includegraphics[width=\cyclgeganimgwidth\linewidth]{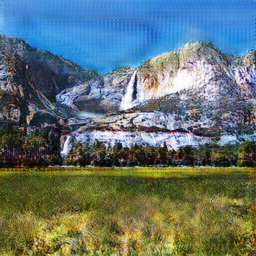} &
		\includegraphics[width=\cyclgeganimgwidth\linewidth]{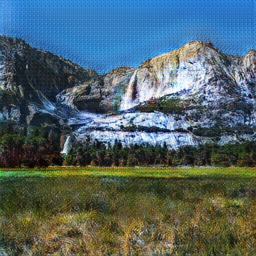} &
		\includegraphics[width=\cyclgeganimgwidth\linewidth]{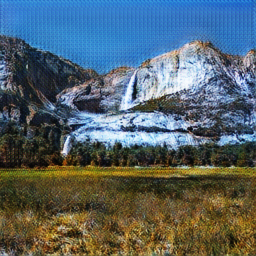}
		\\
		\includegraphics[width=\cyclgeganimgwidth\linewidth]{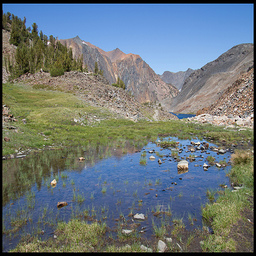} &
		\includegraphics[width=\cyclgeganimgwidth\linewidth]{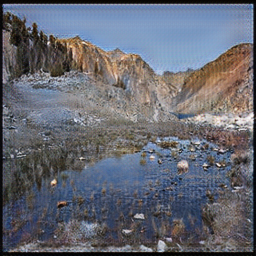} & 
		\includegraphics[width=\cyclgeganimgwidth\linewidth]{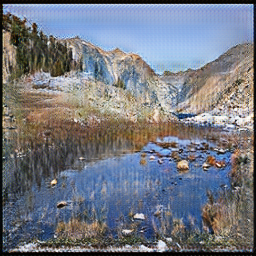} &
		\includegraphics[width=\cyclgeganimgwidth\linewidth]{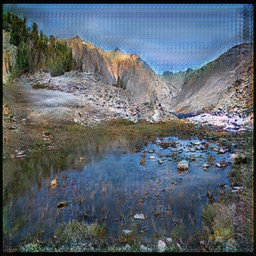} &
		\includegraphics[width=\cyclgeganimgwidth\linewidth]{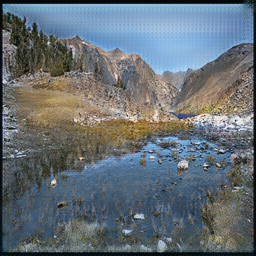} &
		\includegraphics[width=\cyclgeganimgwidth\linewidth]{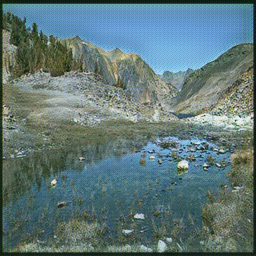} 
	\end{tabular}
	\caption{CycleGAN compression results. Top two rows: horse-to-zebra task. Bottom two rows: summer-to-winter task. Six columns from left to right: source image, style transfer results by original CycleGAN, CEC, GD, GS-32 and GS-8 respectively. FLOPs (in G) and model size (in MB) of each method on each task are annotated above the images.}
	\label{fig:visual_sota_compare}
\end{figure*}

Quantitative comparison results on four different tasks are shown in Table~\ref{tab:quant_sota_compare}.
GS-32 outperforms both CEC and GD on all four tasks, in terms that it achieves better FID (larger $r_f$) with less computational budgets (larger $r_c$ and $r_s$). For example, on horse-to-zebra task, GS-32 has much better FID than both CEC and GD, while achieving more model compactness.
Combined with quantization, our method can further boost the model efficiency (much larger $r_c$) with minimal loss of performance (similar $r_f$). For example, on horse-to-zebra task, GS-8 achieves $4\times$ larger $r_c$ compared with GS-32 with negligible FID drop. On winter-to-summer task, GS-8 compress CycleGAN by $31 \times$ and achieve even slightly better FID.
The visual comparison results are collectively displayed in Fig.~\ref{fig:visual_sota_compare}. We compare the transfer results of four images reported in \cite{shu2019co} for fair comparison. As we can see, the visual quality of GS is better than or at least comparable to those of CEC and GD.

\subsection{Ablation study}

In order to show the superiority of our unified optimization framework over single compression methods and their naive combinations, we conduct thorough ablation studies by comparing the following methods:
\begin{itemize}
    \item Distillation (\ie, GD \cite{chen2020distilling}): Use model distillation alone to train a slim student generator.\footnote{Following \cite{chen2020distilling}, we use student networks with $1/2$ channels of the original generator.} 
    
    \item Channel pruning (CP): Directly use channel pruning during GAN minimax training process. This is implemented by adding $L_{cp}$ to $L_{GAN}$. After channel pruning, we finetune the sub-network by minimax optimizing Eq.~(\ref{eq:GAN}).
    
    \item Cascade: Distillation + CP (D+CP): Use channel pruning to further compress on the student network obtained by model distillation. Then finetune the sub-network by minimax optimizing Eq.~(\ref{eq:GAN}).
    
    \item Cascade: CP + Distillation (CP+D): First do channel pruning on the original network, then use distillation to finetune the pruned network. This method is shown to outperform using channel pruning alone on classification tasks~\cite{wang2018graph}.
    
    \item GS-32: Jointly optimizing channel pruning and distillation.
    
    \item Cascade: GS-32 + quantization (postQ): First use GS-32 to compress the original network, then use 8 bit quantization as post processing and also do quantization-aware finetune on the quantized model by solving problem~(\ref{eq:GAN}).
    
    \item GS-8: Jointly optimizing channel pruning, distillation and quantization.
    
    \item GS-8 (MSE): Replace the perceptual loss in GS-8 by MSE loss.
    
    \item GAN compression with fixed discriminator (\ie, CEC \cite{shu2019co}): Co-evolution based channel pruning. Modeling GAN compression as dense prediction process instead of minimax problem by fixing the discriminator (both network structure and parameter values) during compression process.
\end{itemize}

\begin{figure}[ht]
    \centering
    \begin{subfigure}[t]{0.49\linewidth}
        \centering
        \includegraphics[width=\linewidth]{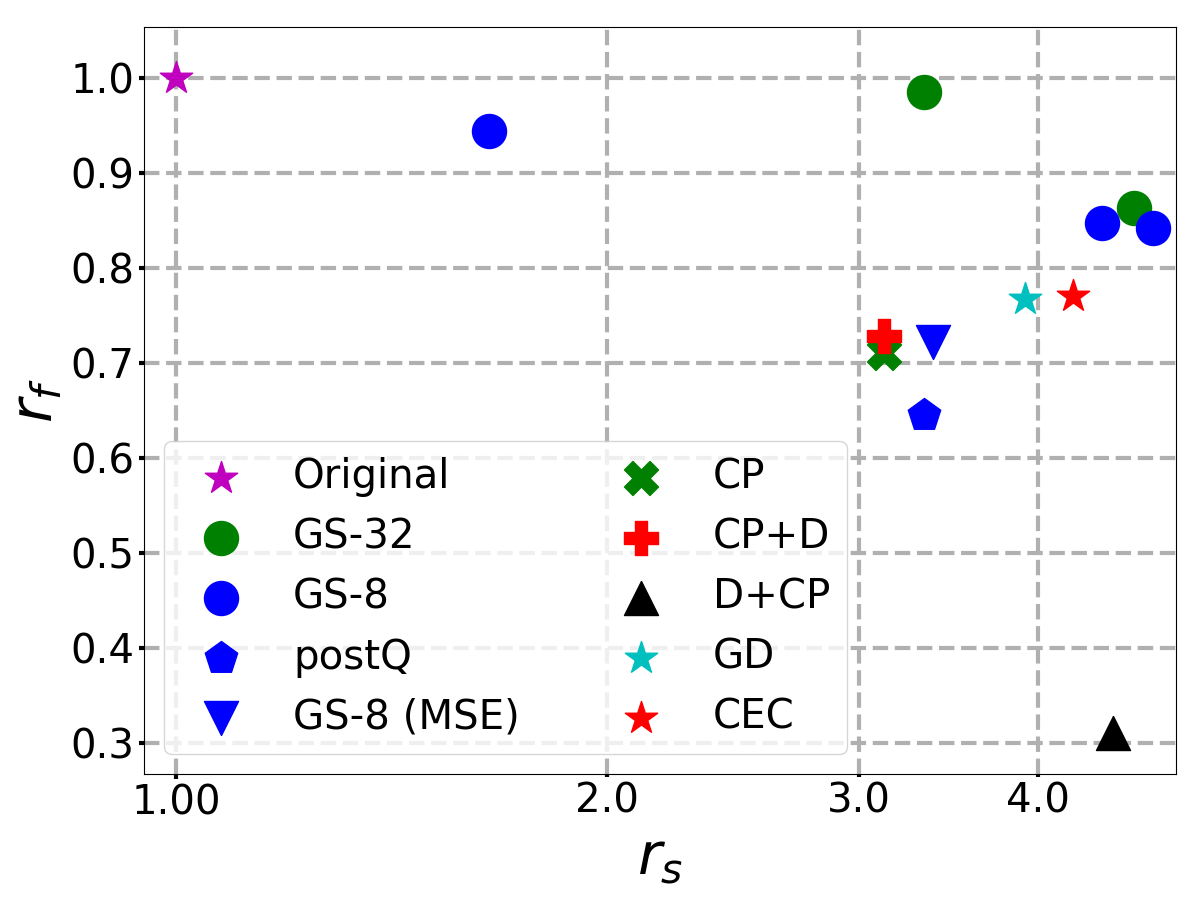}
        \caption{horse-to-zebra task}
    \end{subfigure}%
    ~
    \begin{subfigure}[t]{0.49\linewidth}
        \centering
        \includegraphics[width=\linewidth]{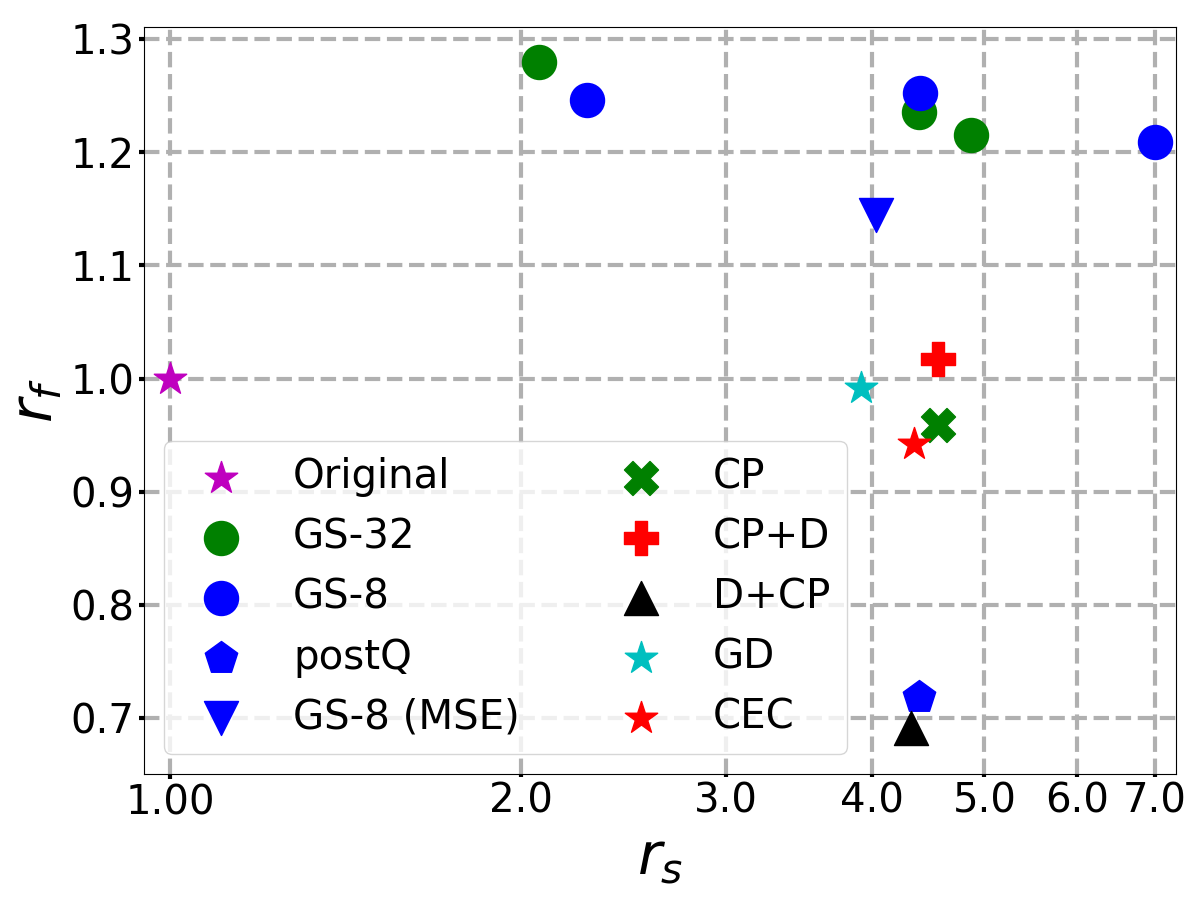}
        \caption{zebra-to-horse task}
    \end{subfigure}
    \caption{Numerical results of ablation studies on horse2zebra dataset.}
    \label{fig:abla_numerical}
\end{figure}

\begin{figure*}[ht] 
	\centering
	\setlength{\tabcolsep}{1pt}
	\begin{tabular}{ccccc}
		Source image & \makecell{Original~\cite{zhu2017unpaired}\\52.90 GFLOPs\\43.51 MB}& \makecell{GD \cite{chen2020distilling}\\13.51 GFLOPs\\10.86 MB} & \makecell{D+CP\\{11.71} GFLOPs\\8.92 MB} & \makecell{CP\\16.95 GFLOPs\\11.07 MB} 
		\\
        \includegraphics[width=\ablaimgwidth\linewidth]{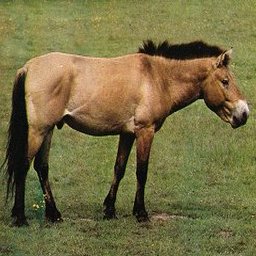} & 
        \includegraphics[width=\ablaimgwidth\linewidth]{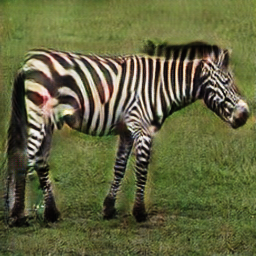} & 
		\includegraphics[width=\ablaimgwidth\linewidth]{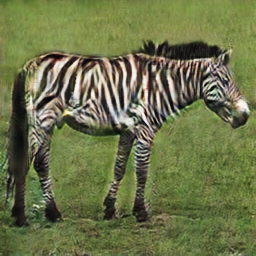} & 
		\includegraphics[width=\ablaimgwidth\linewidth]{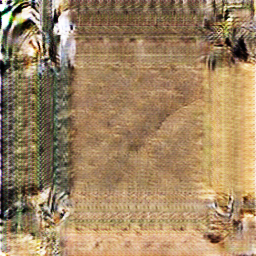} & 
		\includegraphics[width=\ablaimgwidth\linewidth]{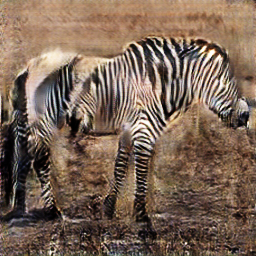} 
		\\
		\makecell{CP+D\\16.95 GFLOPs\\11.07 MB} & \makecell{postQ\\15.90 GFLOPs\\3.13 MB} & \makecell{GS-32\\11.34 GFLOPs\\8.61 MB} & \makecell{GS-8\\10.99 GFLOPs\\2.00 MB} & \makecell{GS-8 (MSE)\\15.66 GFLOPs\\3.22 MB}
		\\
		\includegraphics[width=\ablaimgwidth\linewidth]{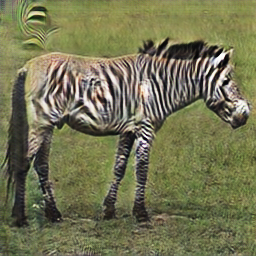} &
		\includegraphics[width=\ablaimgwidth\linewidth]{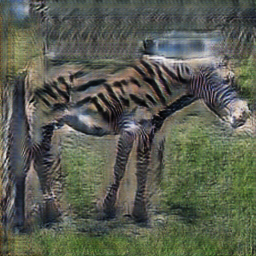} &
        \includegraphics[width=\ablaimgwidth\linewidth]{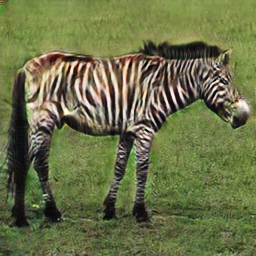} & 
		\includegraphics[width=\ablaimgwidth\linewidth]{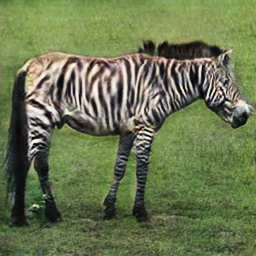} &
		\includegraphics[width=\ablaimgwidth\linewidth]{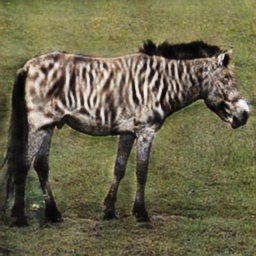} 
	\end{tabular}
	\caption{Visualization results of ablation studies on horse2zebra dataset. FLOPs and model size of each method are annotated above the images.}
	\label{fig:abla_visual}
\end{figure*}

Numerical and visualization results on horse2zebra dataset are shown in Fig.~\ref{fig:abla_numerical} and Fig.~\ref{fig:abla_visual} respectively. As we can see, our unified optimization method achieves superior trade-off between style transfer quality and model efficiency compared with single compression techniques used separately (\eg, CP, GD) and their naive combinations (\eg, CP+D, D+CP, postQ), showing the effectiveness of our unified optimization framework.
For example, directly injecting channel sparsity in GAN minimax optimization (CP) greatly increases the training instability and achieves degraded image generation quality as shown in Fig.~\ref{fig:abla_visual}. This aligns with the conclusions in \cite{shu2019co} that model compression methods developed for classifiers are not directly applicable on GAN compression tasks.
Using model distillation to finetune channel pruned models (CP+D) can indeed improve image generation quality compared with CP, however the generation quality is still much more inferior to our methods at similar compression ratio, as shown in Fig.~\ref{fig:abla_numerical} and Fig.~\ref{fig:abla_visual}. 
Compared with GD, which uses a hand-crafted student network, GS-32 achieves much better FID with even considerably larger compression ratio, showing the effectiveness of jointly searching slim student network structures by channel pruning and training the student network with model distillation.
In contrast, directly using channel pruning to further compress the student generator trained by GD (D+CP) will catastrophically hurt the image translation performance.
Doing post quantization and quantization-aware finetune (postQ) on GS-32 models also suffers great degradation in image translation quality compared with GS-8, showing the necessity to jointly train quantization with channel pruning and model distillation in our unified optimization framework. 
Replacing perceptual loss with MSE loss as $d$ in Eq.~(\ref{eq:distillation}) fails to generate satisfying target images, since MSE loss cannot effectively capture the high-level semantic differences between images.
Last but not least, GS-32 largely outperforms CEC, verifying the effectiveness of incorporating minimax objective into GAN compression problem. 

\subsection{Real-world Application: CartoonGAN}
\label{sec:CartoonGAN}
Finally, we apply GS to a recently proposed style transfer network CartoonGAN, which transforms photos to cartoon images, in order to deploy the model on mobile devices. CartoonGAN has its heavily parameterized generator (56.46 GFLOPs on $256 \times 256$ images) publicly available.\footnote{Available at https://github.com/maciej3031/comixify.} 
Since CartoonGAN has a feed-forward encoder-decoder structure, without using cycle consistent loss, CEC \cite{shu2019co} is not directly applicable to compress it. So we only compare GS with the other published state-of-the-art method GD \cite{chen2020distilling} on this task. Experiments are conducted on the CelebA dataset~\cite{liu2015faceattributes}. 
Following \cite{chen2020distilling}, we use a student generator with $1/6$ channels of the teacher generator for GD, which achieves similar (but less) compression ratio compared with GS.
% We set the initial learning rate $\alpha^{(1)}$ to be $1e^{-3}$ and trained 180 epochs.

% \begin{table}[ht] 
% \begin{center}
% \begin{tabular}{c|c|c}
% \hline
% Model & GFLOPs & Model Size (MB) \\
% \hline\hline
% Original & 56.46 & 42.36 \\
% GD & & \\
% GS-32 & 1.34 & 0.40 \\
% GS-8 & 1.20 & 0.05 \\
% \hline
% \end{tabular}
% \end{center}
% \caption{Statistics of CartoonGAN models.}
% \label{tab:CartoonGAN}
% \end{table}

\begin{figure}[ht] 
	\centering
	\setlength{\tabcolsep}{1pt}
	\begin{tabular}{ccccc}
    	\small Photo images 
		& \small \makecell{Original\\CartoonGAN} 
		& \small \makecell{GD} 
		& \small \makecell{GS-32}
		& \small \makecell{GS-8} 
		\\
		& \small \makecell{56.46 G\\42.34 MB} 
		& \small \makecell{1.41 G \\ 1.04 MB}
		& \small \makecell{1.34 G \\ \cartoondensesize~MB}
		& \small \makecell{1.20 G \\ \cartoonsize~MB} 
		\\

%         \includegraphics[width=\cartoonimgwidth\linewidth]{fig/CartoonGAN/ori_imgs/4_0.png} & 
% 		\includegraphics[width=\cartoonimgwidth\linewidth]{fig/CartoonGAN/fake_gnt_cartoon_color_transfer/4_0.png} & 
% 		\includegraphics[width=\cartoonimgwidth\linewidth]{fig/CartoonGAN/GD_cartoongan/4_0.png} & 
% 		\includegraphics[width=\cartoonimgwidth\linewidth]{fig/CartoonGAN/style_vgg_prune_CartoonGAN_lr0.001_wd0.001_rho0.01_relu1_2_gen_img_color_transfer/4_0.png} &
% 		\includegraphics[width=\cartoonimgwidth\linewidth]{fig/CartoonGAN/GWhat_gen_img_color_transfer/4_0.png} 
% 		\\

%         \includegraphics[width=\cartoonimgwidth\linewidth]{fig/CartoonGAN/ori_imgs/1_4.png} & 
% 		\includegraphics[width=\cartoonimgwidth\linewidth]{fig/CartoonGAN/fake_gnt_cartoon_color_transfer/1_4.png} & 
% 		\includegraphics[width=\cartoonimgwidth\linewidth]{fig/CartoonGAN/GD_cartoongan/1_4.png} & 
% 		\includegraphics[width=\cartoonimgwidth\linewidth]{fig/CartoonGAN/GWhat_gen_img_color_transfer_noisy/1_4.png} &
% 		\includegraphics[width=\cartoonimgwidth\linewidth]{fig/CartoonGAN/GWhat_gen_img_color_transfer/1_4.png} 
% 		\\
		
        \includegraphics[width=\cartoonimgwidth\linewidth]{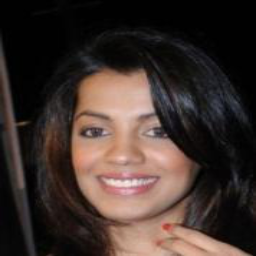} & 
		\includegraphics[width=\cartoonimgwidth\linewidth]{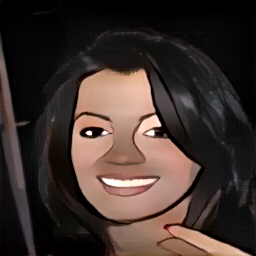} & 
		\includegraphics[width=\cartoonimgwidth\linewidth]{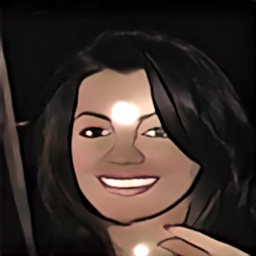} & 
		\includegraphics[width=\cartoonimgwidth\linewidth]{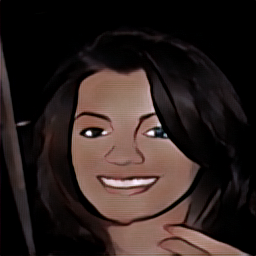} &
		\includegraphics[width=\cartoonimgwidth\linewidth]{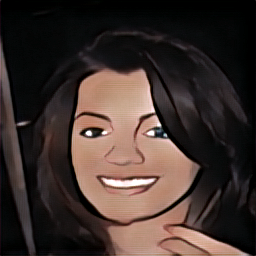} 
		\\

% 		\includegraphics[width=\cartoonimgwidth\linewidth]{fig/CartoonGAN/ori_imgs/6_0.png} & 
% 		\includegraphics[width=\cartoonimgwidth\linewidth]{fig/CartoonGAN/fake_gnt_cartoon_color_transfer/6_0.png} & 
% 		\includegraphics[width=\cartoonimgwidth\linewidth]{fig/CartoonGAN/GD_cartoongan/6_0.png} & 
% 		\includegraphics[width=\cartoonimgwidth\linewidth]{fig/CartoonGAN/style_vgg_prune_CartoonGAN_lr0.001_wd0.001_rho0.01_relu1_2_gen_img_color_transfer/6_0.png} &
% 		\includegraphics[width=\cartoonimgwidth\linewidth]{fig/CartoonGAN/GWhat_gen_img_color_transfer/6_0.png} 
% 		\\
		
% 		\includegraphics[width=\cartoonimgwidth\linewidth]{fig/CartoonGAN/ori_imgs/6_2.png} & 
% 		\includegraphics[width=\cartoonimgwidth\linewidth]{fig/CartoonGAN/fake_gnt_cartoon_color_transfer/6_2.png} & 
% 		\includegraphics[width=\cartoonimgwidth\linewidth]{fig/CartoonGAN/GD_cartoongan/6_2.png} & 
% 		\includegraphics[width=\cartoonimgwidth\linewidth]{fig/CartoonGAN/style_vgg_prune_CartoonGAN_lr0.001_wd0.001_rho0.01_relu1_2_gen_img_color_transfer/6_2.png} &
% 		\includegraphics[width=\cartoonimgwidth\linewidth]{fig/CartoonGAN/GWhat_gen_img_color_transfer/6_2.png} 
% 		\\
		
		\includegraphics[width=\cartoonimgwidth\linewidth]{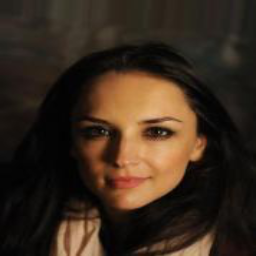} & 
		\includegraphics[width=\cartoonimgwidth\linewidth]{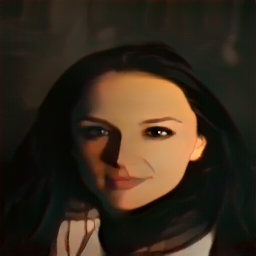} & 
		\includegraphics[width=\cartoonimgwidth\linewidth]{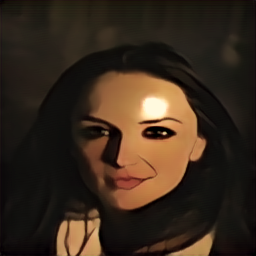} & 
		\includegraphics[width=\cartoonimgwidth\linewidth]{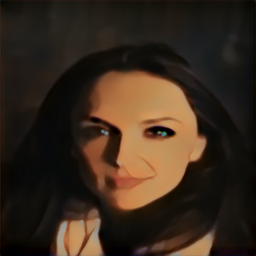} &
		\includegraphics[width=\cartoonimgwidth\linewidth]{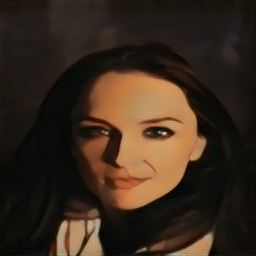} 
		\\
		
% 		\includegraphics[width=\cartoonimgwidth\linewidth]{fig/CartoonGAN/ori_imgs/2_3.png} & 
% 		\includegraphics[width=\cartoonimgwidth\linewidth]{fig/CartoonGAN/fake_gnt_cartoon_color_transfer/2_3.png} &
% 		\includegraphics[width=\cartoonimgwidth\linewidth]{fig/CartoonGAN/GD_cartoongan/2_3.png} &
% 		\includegraphics[width=\cartoonimgwidth\linewidth]{fig/CartoonGAN/style_vgg_prune_CartoonGAN_lr0.001_wd0.001_rho0.01_relu1_2_gen_img_color_transfer/2_3.png} &
% 		\includegraphics[width=\cartoonimgwidth\linewidth]{fig/CartoonGAN/GWhat_gen_img_color_transfer/2_3.png} 
% 		\\

		\includegraphics[width=\cartoonimgwidth\linewidth]{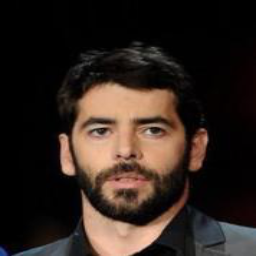} & 
		\includegraphics[width=\cartoonimgwidth\linewidth]{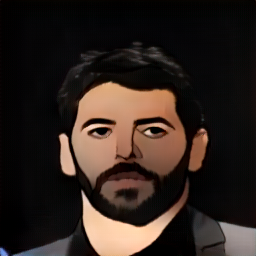} & 
		\includegraphics[width=\cartoonimgwidth\linewidth]{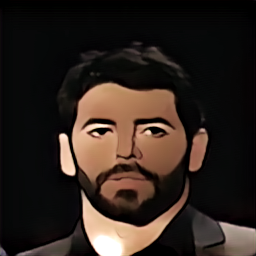} & 
		\includegraphics[width=\cartoonimgwidth\linewidth]{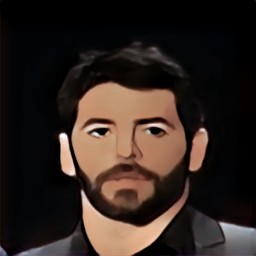} &
		\includegraphics[width=\cartoonimgwidth\linewidth]{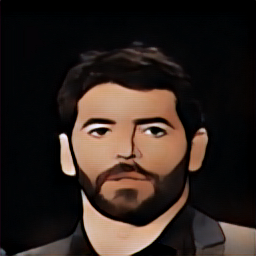} 
		\\
		
		\includegraphics[width=\cartoonimgwidth\linewidth]{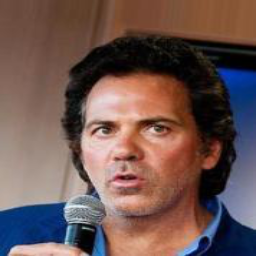} & 
		\includegraphics[width=\cartoonimgwidth\linewidth]{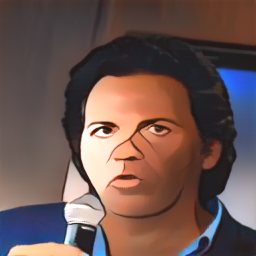} & 
		\includegraphics[width=\cartoonimgwidth\linewidth]{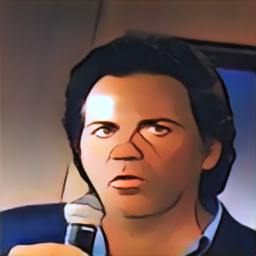} & 
		\includegraphics[width=\cartoonimgwidth\linewidth]{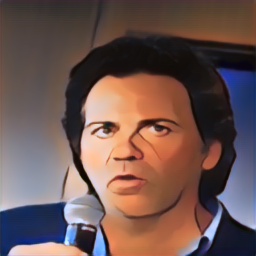} &
		\includegraphics[width=\cartoonimgwidth\linewidth]{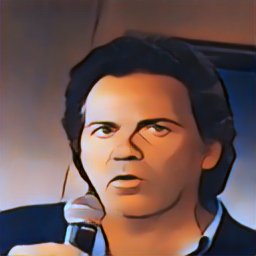} 
		\\
        
	\end{tabular}
	\caption{CartoonGAN compression results. From left to right columns: original photo images, cartoon images generated by original CartoonGAN, GD \cite{chen2020distilling}, GS-32 and GS-8 compressed models, respectively. Corresponding FLOPs (in G) and model size (in MB) are annotated on top of each column.}
	\label{fig:CartoonGAN}
\end{figure}

The visual results of cartoon style transfer, together with model statistics (FLOPs and model sizes), are shown in Fig.~\ref{fig:CartoonGAN}.\footnote{Following \cite{gatys2016preserving}, we use color matching as the post-processing on all compared methods, for better visual display quality.}
All FLOPs are calculated for input images with shape $256 \times 256$. At large compression ratio, the style transfer results of GD have obvious visual artifacts (\eg, abnormal white spots).
In contrast, GS-32 can remarkably compress the original generator by around $\cartoondenseratio \times$ (in terms of FLOPs) with minimal degradation in the visual quality. 
GS-8 can further improve the FLOPs compression ratio to $\cartoonratio \times$ with almost identical visual quality. 
These results again show the superiority of our student generator jointly learned by channel pruning, quantization and distillation, over the hand-crafted student generator used in GD.
Part of the proposed GS framework is integrated into some style transfer products in Kwai Inc.'s Apps. 

\section{Conclusion}

In this paper, we propose the first end-to-end optimization framework combining multiple compression techniques for GAN compression. 
Our method integrates model distillation, channel pruning and quantization, within one unified minimax optimization framework. 
Experimental results show that our method largely outperforms existing GAN compression options which utilize single compression techniques. 
% Specifically, our method compresses CartoonGAN, a state-of-the-art style transfer network, by up to $\mathbf{\cartoonratio \times}$ times, with minimal visual quality degradation.
Detailed ablation studies show that naively stacking different compression methods fails to achieve satisfying GAN compression results, sometimes even hurting the performance catastrophically, therefore testifying the necessity of our unified optimization framework.

\clearpage

\bibliographystyle{splncs04}
\bibliography{reference}

\begin{thebibliography}{10}
\providecommand{\url}[1]{\texttt{#1}}
\providecommand{\urlprefix}{URL }
\providecommand{\doi}[1]{https://doi.org/#1}

\bibitem{bulo2016dropout}
Bul{\`o}, S.R., Porzi, L., Kontschieder, P.: Dropout distillation. In:
  International Conference on Machine Learning. pp. 99--107 (2016)

\bibitem{chen2020frequency}
Chen, H., Wang, Y., Shu, H., Tang, Y., Xu, C., Shi, B., Xu, C., Tian, Q., Xu,
  C.: Frequency domain compact {3D} convolutional neural networks. In: IEEE
  Conference on Computer Vision and Pattern Recognition. pp. 1641--1650 (2020)

\bibitem{chen2020distilling}
Chen, H., Wang, Y., Shu, H., Wen, C., Xu, C., Shi, B., Xu, C., Xu, C.:
  Distilling portable generative adversarial networks for image translation.
  In: AAAI Conference on Artificial Intelligence (2020)

\bibitem{chen2020learning}
Chen, H., Wang, Y., Xu, C., Xu, C., Tao, D.: Learning student networks via
  feature embedding. IEEE Transactions on Neural Networks and Learning Systems
  (2020)

\bibitem{chen2019data}
Chen, H., Wang, Y., Xu, C., Yang, Z., Liu, C., Shi, B., Xu, C., Xu, C., Tian,
  Q.: Data-free learning of student networks. In: IEEE International Conference
  on Computer Vision. pp. 3514--3522 (2019)

\bibitem{chen2020addernet}
Chen, H., Wang, Y., Xu, C., Shi, B., Xu, C., Tian, Q., Xu, C.: {AdderNet}: Do
  we really need multiplications in deep learning? In: IEEE Conference on
  Computer Vision and Pattern Recognition. pp. 1468--1477 (2020)

\bibitem{chen2018cartoongan}
Chen, Y., Lai, Y.K., Liu, Y.J.: {CartoonGAN}: Generative adversarial networks
  for photo cartoonization. In: IEEE Conference on Computer Vision and Pattern
  Recognition. pp. 9465--9474 (2018)

\bibitem{courbariaux2015binaryconnect}
Courbariaux, M., Bengio, Y., David, J.P.: {BinaryConnect}: Training deep neural
  networks with binary weights during propagations. In: Advances in Neural
  Information Processing Systems. pp. 3123--3131 (2015)

\bibitem{fu2020autogan}
Fu, Y., Chen, W., Wang, H., Li, H., Lin, Y., Wang, Z.: {AutoGAN-Distiller}:
  Searching to compress generative adversarial networks. In: International
  Conference on Machine Learning (2020)

\bibitem{gatys2016preserving}
Gatys, L.A., Bethge, M., Hertzmann, A., Shechtman, E.: Preserving color in
  neural artistic style transfer. arXiv preprint arXiv:1606.05897  (2016)

\bibitem{gong2019autogan}
Gong, X., Chang, S., Jiang, Y., Wang, Z.: {AutoGAN}: Neural architecture search
  for generative adversarial networks. In: IEEE International Conference on
  Computer Vision. pp. 3224--3234 (2019)

\bibitem{goodfellow2014generative}
Goodfellow, I., Pouget-Abadie, J., Mirza, M., Xu, B., Warde-Farley, D., Ozair,
  S., Courville, A., Bengio, Y.: Generative adversarial nets. In: Advances in
  Neural Information Processing Systems. pp. 2672--2680 (2014)

\bibitem{gui2020review}
Gui, J., Sun, Z., Wen, Y., Tao, D., Ye, J.: A review on generative adversarial
  networks: {Algorithms}, theory, and applications. arXiv preprint
  arXiv:2001.06937  (2020)

\bibitem{gui2019model}
Gui, S., Wang, H., Yang, H., Yu, C., Wang, Z., Liu, J.: Model compression with
  adversarial robustness: {A} unified optimization framework. In: Advances in
  Neural Information Processing Systems. pp. 1283--1294 (2019)

\bibitem{gulrajani2017improved}
Gulrajani, I., Ahmed, F., Arjovsky, M., Dumoulin, V., Courville, A.C.: Improved
  training of {Wasserstein GANs}. In: Advances in Neural Information Processing
  Systems. pp. 5767--5777 (2017)

\bibitem{guo2020positive}
Guo, T., Xu, C., Huang, J., Wang, Y., Shi, B., Xu, C., Tao, D.: On
  positive-unlabeled classification in {GAN}. In: IEEE Conference on Computer
  Vision and Pattern Recognition. pp. 8385--8393 (2020)

\bibitem{han2020ghostnet}
Han, K., Wang, Y., Tian, Q., Guo, J., Xu, C., Xu, C.: {GhostNet}: {More}
  features from cheap operations. In: IEEE Conference on Computer Vision and
  Pattern Recognition. pp. 1580--1589 (2020)

\bibitem{han2015deep}
Han, S., Mao, H., Dally, W.J.: Deep compression: Compressing deep neural
  networks with pruning, trained quantization and {Huffman} coding. arXiv
  preprint arXiv:1510.00149  (2015)

\bibitem{he2018amc}
He, Y., Lin, J., Liu, Z., Wang, H., Li, L.J., Han, S.: {AMC}: {AutoML} for
  model compression and acceleration on mobile devices. In: European Conference
  on Computer Vision. pp. 784--800 (2018)

\bibitem{he2017channel}
He, Y., Zhang, X., Sun, J.: Channel pruning for accelerating very deep neural
  networks. In: IEEE International Conference on Computer Vision. pp.
  1389--1397 (2017)

\bibitem{heusel2017gans}
Heusel, M., Ramsauer, H., Unterthiner, T., Nessler, B., Hochreiter, S.: {GANs}
  trained by a two time-scale update rule converge to a local {Nash}
  equilibrium. In: Advances in Neural Information Processing Systems. pp.
  6626--6637 (2017)

\bibitem{Hinton_2015}
Hinton, G., Vinyals, O., Dean, J.: Distilling the knowledge in a neural
  network. arXiv preprint arXiv:1503.02531  (2015)

\bibitem{hu2016network}
Hu, H., Peng, R., Tai, Y.W., Tang, C.K.: Network trimming: A data-driven neuron
  pruning approach towards efficient deep architectures. arXiv preprint
  arXiv:1607.03250  (2016)

\bibitem{huang2018data}
Huang, Z., Wang, N.: Data-driven sparse structure selection for deep neural
  networks. In: European Conference on Computer Vision. pp. 304--320 (2018)

\bibitem{hubara2017quantized}
Hubara, I., Courbariaux, M., Soudry, D., El-Yaniv, R., Bengio, Y.: Quantized
  neural networks: {Training} neural networks with low precision weights and
  activations. Journal of Machine Learning Research  \textbf{18}(1),
  6869--6898 (2017)

\bibitem{jiang2019enlightengan}
Jiang, Y., Gong, X., Liu, D., Cheng, Y., Fang, C., Shen, X., Yang, J., Zhou,
  P., Wang, Z.: {EnlightenGAN}: Deep light enhancement without paired
  supervision. arXiv preprint arXiv:1906.06972  (2019)

\bibitem{johnson2016perceptual}
Johnson, J., Alahi, A., Li, F.F.: Perceptual losses for real-time style
  transfer and super-resolution. In: European Conference on Computer Vision.
  pp. 694--711 (2016)

\bibitem{jung2019learning}
Jung, S., Son, C., Lee, S., Son, J., Han, J.J., Kwak, Y., Hwang, S.J., Choi,
  C.: Learning to quantize deep networks by optimizing quantization intervals
  with task loss. In: IEEE Conference on Computer Vision and Pattern
  Recognition. pp. 4350--4359 (2019)

\bibitem{karras2017progressive}
Karras, T., Aila, T., Laine, S., Lehtinen, J.: Progressive growing of {GANs}
  for improved quality, stability, and variation. In: International Conference
  on Learning Representations (2018)

\bibitem{karras2019style}
Karras, T., Laine, S., Aila, T.: A style-based generator architecture for
  generative adversarial networks. In: IEEE Conference on Computer Vision and
  Pattern Recognition. pp. 4401--4410 (2019)

\bibitem{kim2010tree}
Kim, S., P~Xing, E.: Tree-guided group lasso for multi-task regression with
  structured sparsity. The Annals of Applied Statistics  \textbf{6}(3),
  1095--1117 (2012)

\bibitem{kingma2014adam}
Kingma, D.P., Ba, J.: Adam: A method for stochastic optimization. arXiv
  preprint arXiv:1412.6980  (2014)

\bibitem{kupyn2019deblurgan}
Kupyn, O., Martyniuk, T., Wu, J., Wang, Z.: {DeblurGAN-v2}: Deblurring
  (orders-of-magnitude) faster and better. In: IEEE International Conference on
  Computer Vision. pp. 8878--8887 (2019)

\bibitem{ledig2017photo}
Ledig, C., Theis, L., Husz{\'a}r, F., Caballero, J., Cunningham, A., Acosta,
  A., Aitken, A., Tejani, A., Totz, J., Wang, Z., Shi, W.: Photo-realistic
  single image super-resolution using a generative adversarial network. In:
  IEEE Conference on Computer Vision and Pattern Recognition. pp. 4681--4690
  (2017)

\bibitem{li2020gan}
Li, M., Lin, J., Ding, Y., Liu, Z., Zhu, J.Y., Han, S.: {GAN} compression:
  Efficient architectures for interactive conditional {GANs}. In: IEEE
  Conference on Computer Vision and Pattern Recognition. pp. 5284--5294 (2020)

\bibitem{lin2017runtime}
Lin, J., Rao, Y., Lu, J., Zhou, J.: Runtime neural pruning. In: Advances in
  Neural Information Processing Systems. pp. 2181--2191 (2017)

\bibitem{liu2020adadeep}
Liu, S., Du, J., Nan, K., Wang, A., Lin, Y., et~al.: Adadeep: A usage-driven,
  automated deep model compression framework for enabling ubiquitous
  intelligent mobiles. arXiv preprint arXiv:2006.04432  (2020)

\bibitem{liu2017learning}
Liu, Z., Li, J., Shen, Z., Huang, G., Yan, S., Zhang, C.: Learning efficient
  convolutional networks through network slimming. In: IEEE International
  Conference on Computer Vision. pp. 2736--2744 (2017)

\bibitem{liu2015faceattributes}
Liu, Z., Luo, P., Wang, X., Tang, X.: Deep learning face attributes in the
  wild. In: IEEE International Conference on Computer Vision. pp. 3730--3738
  (2015)

\bibitem{lopez_2015}
Lopez-Paz, D., Bottou, L., Sch{\"o}lkopf, B., Vapnik, V.: Unifying distillation
  and privileged information. arXiv preprint arXiv:1511.03643  (2015)

\bibitem{luo2017thinet}
Luo, J.H., Wu, J., Lin, W.: {ThiNet}: A filter level pruning method for deep
  neural network compression. In: IEEE International Conference on Computer
  Vision. pp. 5058--5066 (2017)

\bibitem{mishra2017apprentice}
Mishra, A., Marr, D.: Apprentice: {Using} knowledge distillation techniques to
  improve low-precision network accuracy. arXiv preprint arXiv:1711.05852
  (2017)

\bibitem{miyato2018spectral}
Miyato, T., Kataoka, T., Koyama, M., Yoshida, Y.: Spectral normalization for
  generative adversarial networks. In: International Conference on Learning
  Representations (2018)

\bibitem{molchanov2019importance}
Molchanov, P., Mallya, A., Tyree, S., Frosio, I., Kautz, J.: Importance
  estimation for neural network pruning. In: IEEE Conference on Computer Vision
  and Pattern Recognition. pp. 11264--11272 (2019)

\bibitem{parikh2014proximal}
Parikh, N., Boyd, S., et~al.: Proximal algorithms. Foundations and Trends in
  Optimization  \textbf{1}(3),  127--239 (2014)

\bibitem{polino2018model}
Polino, A., Pascanu, R., Alistarh, D.: Model compression via distillation and
  quantization. arXiv preprint arXiv:1802.05668  (2018)

\bibitem{rastegari2016xnor}
Rastegari, M., Ordonez, V., Redmon, J., Farhadi, A.: {XNOR-Net}: {ImageNet}
  classification using binary convolutional neural networks. In: European
  Conference on Computer Vision. pp. 525--542 (2016)

\bibitem{sanakoyeu2018style}
Sanakoyeu, A., Kotovenko, D., Lang, S., Ommer, B.: A style-aware content loss
  for real-time {HD} style transfer. In: European Conference on Computer
  Vision. pp. 698--714 (2018)

\bibitem{shen2019searching}
Shen, M., Han, K., Xu, C., Wang, Y.: Searching for accurate binary neural
  architectures. In: IEEE International Conference on Computer Vision Workshops
  (2019)

\bibitem{shu2019co}
Shu, H., Wang, Y., Jia, X., Han, K., Chen, H., Xu, C., Tian, Q., Xu, C.:
  {Co-Evolutionary} compression for unpaired image translation. In: IEEE
  International Conference on Computer Vision. pp. 3235--3244 (2019)

\bibitem{singh2020leveraging}
Singh, P., Verma, V.K., Rai, P., Namboodiri, V.: Leveraging filter correlations
  for deep model compression. In: IEEE Winter Conference on Applications of
  Computer Vision. pp. 835--844 (2020)

\bibitem{theis2018faster}
Theis, L., Korshunova, I., Tejani, A., Husz{\'a}r, F.: Faster gaze prediction
  with dense networks and fisher pruning. arXiv preprint arXiv:1801.05787
  (2018)

\bibitem{tung2018clip}
Tung, F., Mori, G.: {CLIP-Q}: {Deep} network compression learning by
  in-parallel pruning-quantization. In: IEEE Conference on Computer Vision and
  Pattern Recognition. pp. 7873--7882 (2018)

\bibitem{wang2019haq}
Wang, K., Liu, Z., Lin, Y., Lin, J., Han, S.: {HAQ}: {Hardware-Aware} automated
  quantization with mixed precision. In: IEEE Conference on Computer Vision and
  Pattern Recognition. pp. 8612--8620 (2019)

\bibitem{wang2018graph}
Wang, M., Zhang, Q., Yang, J., Cui, X., Lin, W.: Graph-adaptive pruning for
  efficient inference of convolutional neural networks. arXiv preprint
  arXiv:1811.08589  (2018)

\bibitem{wang2018learning}
Wang, Y., Xu, C., Chunjing, X., Xu, C., Tao, D.: Learning versatile filters for
  efficient convolutional neural networks. In: Advances in Neural Information
  Processing Systems. pp. 1608--1618 (2018)

\bibitem{wang2018adversarial}
Wang, Y., Xu, C., Xu, C., Tao, D.: Adversarial learning of portable student
  networks. In: AAAI Conference on Artificial Intelligence. pp. 4260--4267
  (2018)

\bibitem{wang2018packing}
Wang, Y., Xu, C., Xu, C., Tao, D.: Packing convolutional neural networks in the
  frequency domain. IEEE Transactions on Pattern Analysis and Machine
  Intelligence  \textbf{41}(10),  2495--2510 (2018)

\bibitem{wen2016learning}
Wen, W., Wu, C., Wang, Y., Chen, Y., Li, H.: Learning structured sparsity in
  deep neural networks. In: Advances in Neural Information Processing Systems.
  pp. 2074--2082 (2016)

\bibitem{wu2018deep}
Wu, J., Wang, Y., Wu, Z., Wang, Z., Veeraraghavan, A., Lin, Y.:
  {Deep-$k$-Means}: Re-training and parameter sharing with harder cluster
  assignments for compressing deep convolutions. In: International Conference
  on Machine Learning. pp. 5363--5372 (2018)

\bibitem{yang2019ecc}
Yang, H., Zhu, Y., Liu, J.: {ECC}: {Platform-independent} energy-constrained
  deep neural network compression via a bilinear regression model. In: IEEE
  Conference on Computer Vision and Pattern Recognition. pp. 11206--11215
  (2019)

\bibitem{yang2019controllable}
Yang, S., Wang, Z., Wang, Z., Xu, N., Liu, J., Guo, Z.: Controllable artistic
  text style transfer via shape-matching {GAN}. In: IEEE International
  Conference on Computer Vision. pp. 4442--4451 (2019)

\bibitem{zhao2020smartexchange}
Zhao, Y., Chen, X., Wang, Y., Li, C., You, H., Fu, Y., Xie, Y., Wang, Z., Lin,
  Y.: Smartexchange: Trading higher-cost memory storage/access for lower-cost
  computation. arXiv preprint arXiv:2005.03403  (2020)

\bibitem{zhu2016trained}
Zhu, C., Han, S., Mao, H., Dally, W.J.: Trained ternary quantization. arXiv
  preprint arXiv:1612.01064  (2016)

\bibitem{zhu2017unpaired}
Zhu, J.Y., Park, T., Isola, P., Efros, A.A.: Unpaired image-to-image
  translation using cycle-consistent adversarial networks. In: IEEE
  International Conference on Computer Vision. pp. 2223--2232 (2017)

\end{thebibliography}

\clearpage
\appendix
\section{Image Generation with SNGAN} \label{sec:sngan}

\begin{figure*}[ht]
	\centering
	\begin{tabular}{ccc}
    	Original SNGAN & GS-8 & GS-8
    	\\
        \includegraphics[width=0.26\linewidth]{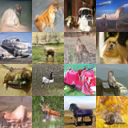} &
    	\includegraphics[width=0.26\linewidth]{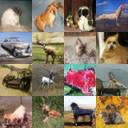} & 
    	\includegraphics[width=0.26\linewidth]{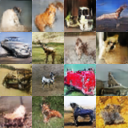} 
    	\\
    	\makecell{1602.75 MFLOPs\\(IS=8.27)} &
       	\makecell{1115.11 MFLOPs\\(IS=8.14)} &
       	\makecell{510.33 MFLOPs\\(IS=7.62)}
       	% \makecell{292.22 MFLOPs\\(IS=7.13)} 
	\end{tabular}
	\caption{CIFAR-10 images generation by SNGAN (original and compressed). Leftmost column: images generated by original SNGAN. The rest columns: images generated by GS-8 compressed SNGAN, with different compression ratios. 
	Images are randomly selected instead of cherry-picked.}
    \label{fig:sngan}
\end{figure*}

We have demonstrated the effectiveness of GS in compressing image-to-image GANs (\eg, CycleGAN~\cite{zhu2017unpaired}, StyleGAN~\cite{sanakoyeu2018style}) in the main text. Here we show GS is also generally applicable to noise-to-image GANs (\eg, SNGAN \cite{miyato2018spectral}).
SNGAN with the ResNet backbone is one of the most popular noise-to-image GANs, with state-of-the-art performance on a few datasets such as CIFAR10 \cite{krizhevsky2009learning}. The generator in SNGAN has 7 convolution layers with 1.57 GFLOPs, with $32 \times 32$ image outputs. We evaluate SNGAN generator compression on the CIFAR-10 dataset. Inception Score (IS) \cite{salimans2016improved} is used to measure image generation and style transfer quality. We use latency (FLOPs) and model size to evaluate the network efficiency.
% We set the initial learning rate $\alpha^{(1)}$ in Algorithm~\ref{alg:GS} to be $1\times10^{-3}$ here. Batch size is 32 and the training takes 156,300 iterations. We obtain generators of different model complexity by varying $\rho$ parameter in Eq.~(\ref{eq:overall}), between $1\times10^{-4}$ to $1\times10^{-3}$. 
Quantative and visualization results are shown in Table \ref{tab:sngan} and Figure \ref{fig:sngan} respectively. 
GS is able to compress SNGAN by up to $8 \times$ (in terms of model size), with minimum drop in both visual quality and the quantitative IS value of generated images.

\begin{table}[ht]
\begin{center}
\caption{SNGAN compression results.}
\label{tab:sngan}
\begin{tabular}{c|c|c|c}
\hline
Method & MFLOPs & \makecell{Model Size \\ (MB)} & IS \\
\hline\hline
Original & 1602.75 & 16.28 & 8.27 \\
\hline
\multirow{2}{*}{GS-32} 
& 1108.78 & 12.88 & 8.01 \\
& 509.39 & 8.32 & 7.65 \\
% & {254.28} & 5.08 & 7.30 \\
\hline
\multirow{2}{*}{GS-8} 
& 1115.11 & 3.24 & 8.14	\\	
& 510.33 & 2.01 & 7.62 \\	
% & 292.22 & {1.31} & 7.13 \\	
\hline
\end{tabular}
\end{center}
\end{table}

\end{document}